# Nonlinear Markov Clustering by Minimum Curvilinear Sparse Similarity

C. Durán, A. Acevedo, S. Ciucci, A. Muscoloni, and CV. Cannistraci

**Abstract**— The development of algorithms for unsupervised pattern recognition by nonlinear clustering is a notable problem in data science. Markov clustering (MCL) is a renowned algorithm that simulates stochastic flows on a network of sample similarities to detect the structural organization of clusters in the data, but it has never been generalized to deal with data nonlinearity. Minimum Curvilinearity (MC) is a principle that approximates nonlinear sample distances in the high-dimensional feature space by curvilinear distances, which are computed as transversal paths over their minimum spanning tree, and then stored in a kernel. Here we propose MC-MCL, which is the first nonlinear kernel extension of MCL and exploits Minimum Curvilinearity to enhance the performance of MCL in real and synthetic data with underlying nonlinear patterns. MC-MCL is compared with baseline clustering methods, including DBSCAN, K-means and affinity propagation. We find that Minimum Curvilinearity provides a valuable framework to estimate nonlinear distances also when its kernel is applied in combination with MCL. Indeed, MC-MCL overcomes classical MCL and even baseline clustering algorithms in different nonlinear datasets.

— — — — — — — — — ◆ — — — — — — — — — —

## 1 INTRODUCTION

CLUSTERING is an unsupervised pattern recognition methodology which, given an ensemble of objects or data, aims to recognize their organization in groups and subgroups (with or without hierarchical structure) starting from their features. Clustering can be seen as one of the oldest strategies to understand and to interpret pattern formation in our world: indeed, in their daily life, people expresses their intelligence also in the action to group objects, items or even time-series events in relation to the similarity or dissimilarity between their features [1]. Clustering is used also to detect communities in networks [2]–[4]. Nowadays, in artificial intelligence, clustering is defined as the automatic and unsupervised identification of groups of observations that are similar to one another and different from other groups in a dataset [5]. Indeed, clustering aims mainly to identify distributions and patterns in the underlying data, generating a partitioning of a given dataset into different groups called clusters [6]. In this sense, the patterns of the observations that are grouped in the same cluster should be similar (in the feature space) to each other, while patterns of observations that result in different clusters should not.

In the era of Big Data, there is a tremendous amount of high-dimensional data available due to the progress in storage procedures, and the ubiquitous growth and exploitation of technologies that generate high-dimensional datasets is a trend that will persist in the next decades [7]. However, in fields such as systems biology and molecular medicine, the realization of controlled experiments that are able to provide observations or samples to investigate a scientific hypothesis can be very time-consuming (recruitment of patients, lab experiments etc.) and also expensive [8]. For such a reason, in these fields, pilot studies that generate few amount of samples, in order to test the validity of a scientific hypothesis before to take the decision to scale to the big numbers, is a frequent practice [8]. Under this context, recently, a novel principle called Minimum Curvilinearity (MC) [9] was proposed with the aim to reveal nonlinear patterns in data, especially, in the case of high-dimensional datasets with few samples and many features. MC is a principle that suggests to estimate nonlinear sample distances in the high-dimensional feature space by curvilinear distances, which are computed as transversal paths over their minimum spanning tree, and then stored in a kernel. The MC-kernel has been used in combination with several unsupervised and supervised machine learning techniques to solve nonlinear problems [9]–[12]. In particular, it has been employed as a distance kernel for Affinity Propagation (AP) - a clustering algorithm based on message passing - and their combination consists in a new algorithm for nonlinear clustering which was termed MC-AP [9]. MC-AP was tested on many datasets, such as for instance image proteomic data related to neuropathic pain and, differently from classical AP, it successfully revealed clusters of patients with and without pain.

Here, we focus the present study on a landmark clustering method called Markov Clustering (MCL), which is an unsupervised algorithm for clustering of nodes in weighted graphs and is based on simulations of stochastic

• *C. Duran, A. Acevedo, S. Ciucci, A. Muscoloni are with the Biomedical Cybernetics Group, Biotechnology Center (BIOTEC), Center for Molecular and Cellular Bioengineering (CMCB), Center for Systems Biology Dresden (CSBD), Technische Universität Dresden, Dresden, Germany. E-mail: aldo.acevedo.toledo@gmail.com, claudio134@gmail.com, saraciuccipg@gmail.com, juo.ming.ju@gmail.com.*
• *CV. Cannistraci is with the Biomedical Cybernetics Group, Biotechnology Center (BIOTEC), Center for Molecular and Cellular Bioengineering (CMCB), Center for Systems Biology Dresden (CSBD), Technische Universität Dresden, Dresden, Germany and with the Complex Network Intelligence center, Tsinghua Laboratory of Brain and Intelligence, Tsinghua University, Beijing, China. E-mail: kalokagathos.agon@gmail.com.*

flows [13] (http://micans.org/mcl/). The accepted procedure in data analysis is that MCL receives in input a n × n similarity matrix (n is the number of samples) which resembles the structure of weighted similarity graph, and it is used as a framework to simulate stochastic flows and detect the structural organization of node clusters. The similarity matrix is usually computed as a correlation network (often by means of Pearson correlation) between the samples. Since a similarity can be also estimated as the inversion of a dissimilarity (or distance), in this study we design a new ad-hoc sparse MC-similarity-kernel that contains similarities which are obtained by inverting and automatically pruning the original pairwise MC-distances. This sparse MC-similarity-kernel is used as input for the MCL algorithm and this gives rise to MC-MCL: a new nonlinear clustering algorithm for the analysis of data with underlying nonlinear patterns. In practice, this new algorithm for clustering is a nonlinear and sparse version of the classical MCL, where the nonlinearity is MC-driven and the sparsity is self-learned using the threshold that maximize pruning without losing the one-component similarity network connectivity, which ensures the continuity of the stochastic flows.

We compared the performance of MC-MCL with baseline clustering algorithms such as classical MCL, AP [14], MC-AP [9], density-based spatial clustering of applications with noise (DBSCAN) [15], and K-means [16]. They have been compared both on real and synthetic high-dimensional datasets and using different metrics to evaluate their performances. The results obtained across the different comparisons suggest that MC effectively addresses the problem of data nonlinearity. This improves the performance of stochastic flow clustering to the extent that MC-MCL clearly outperforms classical MCL in all the tests, and even other landmark clustering algorithms in the general evaluation framework.

## 2 Methods

### 2.1 Minimum Curvilinearity

Minimum Curvilinearity (MC) [9] – introduced by Cannistraci et al. in 2010 - was invented with the aim to reveal nonlinear patterns in data, especially in the case of datasets with few samples and many features. Nonlinearity is often driven by hierarchy and - under the hypothesis that at least part of data nonlinearity is associated to a generative process that forces sample hierarchy - the basic idea behind MC is to exploit the hierarchical organization and structure of the samples in the feature space to approximate their pairwise nonlinear relationship. Indeed, the MC principle suggests that nonlinear curvilinear distances between samples can be estimated as transversal paths over their Minimum Spanning Tree (MST), which is constructed according to a certain distance (Euclidean, correlation-based, etc.) in a multidimensional feature space. In this study, we considered Pearson-correlation-based and Euclidean-based distances (refer to [9] for details). The collection of all MC pairwise distances forms a distance matrix called the MC-distance matrix or MC-kernel, which can be used as input in algorithms for dimensionality reduction, clustering, classification and generally in any type of machine learning.

### 2.2 Clustering algorithms

*Markov clustering (MCL)*
MCL [13] – introduced by Stijn van Dongen in 2000 - is an algorithm for data clustering based on simulations of stochastic flows (random walks) in networks, which works with an iterative process by alternating two operators called expansion and inflation. The expansion operator corresponds to the computation of random walks of higher length (many steps), which associates new probabilities between each pair of nodes. In practice, the expansion serves to associate higher probabilities to paths within clusters rather than in between clusters, because in general, there are more ways to go from one node to another in the same cluster. While the inflation operator have the effect of boosting intra-cluster walk probabilities and lowering inter-cluster walks. In practice, the inflation is the MCL parameter that serves to detect clustering patterns on different scales of granularity. The inflation parameter is automatically obtained by binary search, where the search stops when the correct number of clusters are found. Precisely, the value of inflation is searched in the range of [1.1, 20] at different resolutions or steps [0.1, 0.01 and 0.001] to ensure the finding of the correct number of clusters. If the first resolution (0.1) is not enough, the search continues at lower resolution between the last two searched bounds until obtaining the desired number of clusters or arriving to the lowest resolution. The range of search between 1.1 and 20 is defined in order to span a large values interval (compared to the one suggested by the author of the algorithm, which is between 1.1 and 6) that accounts for the different scales of granularity of possible analysed datasets. For clustering samples of a multidimensional dataset, the workflow starts with the computation of similarities (generally Pearson correlations) between the samples, by creating an edge between each pair, where the edge-weight assumes the value of the respective pairwise sample similarity. This produces the weighted similarity network upon which to simulate stochastic flows and detect the structural organization of clusters in the data. We consider in our study two different options to define a similarity: The first option is to use the person correlation. The second option is to adopt Euclidean similarity defined according to this function:

$$ES(x) = (1 - x/\max(x))$$

Where x is a variable that indicates the Euclidean distance between a pair of samples and max(x) is the largest Euclidean distance between all the pairs of samples.

As suggested in the MCL user manual (https://micans.org/mcl/MCL), a network construction and reduction step usually improves the clustering. It means that a sparsification of the weighted similarity matrix - that shapes (construction phase) a network topology by pruning (reduction phase) links with low similarity - is recommended before to start the clustering procedure. As example, the author mention in their user guide to arbitrarily threshold and then discharge similarities lower than 0.7. After, they suggest to rescale the remaining value between [0,0.3]. This should be intended as to rescale between zero and the maximum similarity value in the similarity matrix minus the threshold, because the rescaling ensure stabilities in the stochastic flow clustering procedure. However, there are not indications for a general strategy to follow and, in practice, there is a free parameter to tune for the similarity threshold, and there is not available any automatic procedure. Unlikely, this threshold value should be arbitrarily specified by the user.

*Enforcing network sparsity in Markov clustering (MCL)*
In order to overcome the network threshold issue described at the end of the previous paragraph, we introduce a simple but effective technical innovation to enforce sparsity of the similarity network. In our implementation of the MCL algorithm, we propose a strategy according to which the threshold selection is done automatically by progressively pruning and rescaling the similarity network at increasing similarity threshold values (the unique values of the network weights are ranked and, starting from the lowest value in the list, they are increasingly tested as threshold). The function used for pruning and rescaling is the following:

$$f(x) = ReLU(x - t) = (x - t)^+ = \max[0, (x - t)] \quad (1)$$

Where x is the similarity matrix and t is the threshold (with values including 0 and lower than 1) tested at a certain iteration of the progressive pruning. When the network loses its topological integrity and separates in a number of components larger than one, the procedure stops and this last threshold value is discharged, while the second last threshold value is selected to prune and to rescale the similarity values. In brief, this is a strategy to maximize sparsification of the network topology, while retaining its one-component connectivity.

*Minimum Curvilinear Markov Clustering (MC-MCL)*
With the purpose of creating and testing a nonlinear variant of the MCL algorithm, we propose MC-MCL. The idea is the following: the MC-kernel (refer to the Minimum Curvilinearity section) is a nonlinear kernel that expresses the pairwise relations between samples as a value of distance: a small samples distance indicates high sample similarity, while a large samples distance indicates low sample similarity. As we anticipated in the Minimum Curvilinear section above, in this study we will consider two different distances (Pearson-correlation-based and Euclidean-based) to build the MST, therefore below we will describe respectively two different procedures to derive the MC-similarity kernels. In case the MST and the associated MC-distance kernel are built with Pearson-correlation-distance, we invert the MC-distance kernel to get a MC-similarity kernel and put to zero the negative values (in case of t=0) or all the values lower of a threshold t, using the following function:

$$f(x) = ReLU(1 - x - t) = (1 - x - t)^+ = \max[0, (1 - x - t)] \quad (2)$$

Where: x is the original value of the pairwise MC distance; t is the same threshold defined in equation (1) above to enforce the network sparsity (and it is automatically detected using the same strategy described above); and f(x) is the derived value of the pairwise MC similarity. Therefore, small f(x) values (close to zero) indicate low sample similarity and large f(x) values (close to one) indicate high sample similarity.

Let's clarify now an important property of the MC-similarity defined in (2), and why this inversion is well posed. The MST is computed on a correlation-based distance (cd) that is defined as:

$$0 < cd(y) = (1 - y) < 2, \text{ with } -1 < y < 1 \quad (3)$$

Where y is the original Pearson correlation value and cd → 0 means high similarity, cd=1 means random similarity and cd=2 means anti-similarity (nothing can be more dissimilar than the opposite trend). As a consequence of this mathematical codification of cd, any MC distance that is larger than 1 tends to overcome an intrinsic threshold of random similarity, hence MC distances larger than one can be interpreted as less significant than random. This mechanism, which seems naïve, is in reality refined and allows directly to assess that any MC-distance smaller than 1 is under the natural threshold of random sample similarity association (and should be accepted), therefore any MC-distance larger than 1 can be neglected because is less significant than random similarity. And this is actually what we mathematically define with the ReLU function applied after the 1-x-t inversion in (2). For example: if we fix t=0, a MC-distance x=1.2 is larger than 1 and therefore should be neglected as MC-similarity, indeed f(x) = ReLU(1 - 1.2) = 0. In more general, the equation (2) suggests that we can learn a similarity threshold t=>0 (on the weights of the network) which preserves the network structure and discharge links that are not significant to preserve the integrity of the network flows (because they do not disconnect the network). If t=0, sample similarities (links) that are less significant than random similarities are discharged. If t>0, also sample similarities (links) that are not significant to preserve the stochastic flows are discharged. This naïve strategy allows to induce sparsity in the MC-similarity kernel by means of an intrinsic and self-adaptive thresholding mechanism that neglects connectivity with similarity worse than random and, as a matter of fact, it avoids that the stochastic flows of MCL runs on network branches or zones that would suffer unreliable connectivity.

In case the MST and the associated MC-distance kernel are built with Euclidean-distance, we invert the MC-distance kernel to get a MC-similarity kernel according to the following function:

$$f(x) = ReLU\left(1 - \frac{x}{\max(x)} - t\right) = \left(1 - \frac{x}{\max(x)} - t\right)^+ = \\ \max[0, \left(1 - \frac{x}{\max(x)} - t\right)] \quad (4)$$

Where: x is a variable that indicates the Euclidean-based MC-distance between a pair of samples; max(x) is the largest Euclidean-based MC-distance between all the pairs of samples; and t is the same threshold defined in equation (1) above to enforce the network sparsity (and it is automatically detected using the same strategy described above). A technical detail is that for the computation of the MC-distance kernel (hence before the inversion procedures described above), three alternatives are used: 1) original distances in the MC-kernel (MC-MCLo), 2) their square root $x^{1/2}$ (MC-MCLs), or 3) their logarithm $\log(1+x)$ (MC-MCLl). As already investigated in [9], the square root and the log operators can attenuate the estimation of large distances and, on the contrary, amplifies the estimation of short distances. Consequently, they help to regularize the nonlinear distances inferred over the MST in order to use them for message passing [9] (such as AP) or stochastic flow simulation (such as MCL) clustering algorithms.

The final steps are the same automatic threshold selection described above in order to build the sparse similarity network for the classical MCL, and then to run the standard MCL algorithm on the MC-similarity sparse network (see pseudocode in Table 1 for more details on the MC-MCL computation). In practice, this new algorithm for clustering is a nonlinear and sparse version of the classical MCL, where the nonlinearity is MC-driven and the sparsity is self-learned using the threshold that maximize pruning without losing the one-component similarity network connectivity. The Matlab code with a user guide to implement MC-MCL is available at this link:                    https://github.com/biomedical-cybernetics/minimum-curvilinear-Markov-clustering

### Affinity propagation (AP) and Minimum curvilinear affinity propagation (MC-AP)

AP [14] – introduced by Frey et al. in 2007 - is a clustering algorithm based on a message passing procedure that takes as input a similarity value (in general codified as negative distance/dissimilarity values) between pairs of data points. The messages are propagated between data points until a high-quality set of exemplars and corresponding clusters gradually appear [14]. AP algorithm does not take as input the predefined number of clusters, but requires for each data sample a real number which is termed preference. Samples with larger preferences are more likely to be chosen as exemplars to form a data cluster. The number of identified exemplars (number of clusters) is influenced by the values of the input preferences, but also emerges from the message passing procedure [14]. If a priori, all samples are equally suitable as exemplars, the preferences should be set to a common value. This value can be varied to produce different numbers of clusters. The shared preference value could be the maximum of the input similarities (resulting in a large number of clusters) or their minimum (resulting in a small number of clusters). Here, given in input the expected number of clusters, we implement a binary search that is able to detect the shared preference value that produces a number of clusters as much as closer to the expected one. For reason of space, we refer to the original article of the affinity propagation algorithm in order to check the technical details [14]. Two different distances are considered as inputs: Euclidean and Pearson-correlation-based (refer to [9] for details).

MC-AP [9] - introduced by Cannistraci et al. in 2010 - is also a nonlinear version of AP which is based on the Minimum Curvilinear distance kernel, and it might be a good alternative to AP when the goal is to search for nonlinear patterns in the data. The MC-distance kernel is built according to two different distances as inputs: Euclidean and Pearson-correlation-based (refer to [9] for details). The number of clusters is identified following the same procedure described above for AP.

### Density-based spatial clustering of applications with noise (DBSCAN)

(DBSCAN) [15] - introduced by Ester et al. in 1996 - is a density-based clustering algorithm: given a set of points in a space, it clusters together points that are closely packed (points with many nearby neighbours), denoting as outliers points that lie alone in low-density regions (whose nearest neighbors are too far away). It requires two density parameters: MinPts and Eps. Selected any point *j* in the space, MinPts is the minium number of points inside a neighbourhood (of the selected point *j*) that is defined as a circle of radius Eps. DBSCAN defines *core points* all the points that have at least MinPts points (including itself) in their Eps neighbourhood. The main idea behind this algorithm is that a group of points that are mutually reachable by means of core points (because they are included in the neighbourhood of radius Eps of core points) forms a cluster. All points not reachable from any other point are *outliers* or *noise points*.

This algorithm does not need as input the desired number of clusters, instead it finds them automatically according to the tuning of the two above mentioned parameters. Nevertheless, the finding of these correct parameters MinPts and Eps is a nontrivial problem. MinPts is linearly searched, and for each MinPts, the Eps parameter is explored with a binary search strategy. If the desired number of clusters are found, and there are no noisy points in output (for all datasets, all points belong to a particular cluster), then the search stops. We consider two different distances as inputs: Euclidean and Pearson-correlation-based (refer to [9] for details).

### K-means

K-means [16], [17] – introduced as idea by Steinhaus in 1956 and termed K-means by MacQueen in 1967 – is one of the oldest data clustering algorithm still widely used because it is easy and effective. It splits the data into a set of k desired clusters. It starts with an initial partition of the data and then uses an iterative control strategy to optimize an objective function. Each cluster is represented by the gravity centre of the cluster. In other words, it determines k representatives by minimizing the objective function, then it assigns each sample to the cluster with its closest representative centre. A major restriction is that, generally, the shape of the clusters found by this algorithm is convex (linear data). We consider two different distances as inputs: Euclidean and Pearson-correlation-based (refer to [9] for details).

### 2.3 Procedure to evaluate the performance of clustering algorithms

The clustering algorithms were applied to the 4 datasets described below, either raw, or after a log-based normalization (the function log10(1+x)) was used, where x is original feature value). Their performance was evaluated by means of Accuracy (Acc), a common measure that evaluates the number of correctly predicted labels with respect to the total number of predictions; Adjusted Rand Index (ARI) [18], and Normalized Mutual Information (NMI) [19], [20], which assess the agreement between two partitions, in this case between the true labels of the data and the labels assigned by the clustering algorithm. ARI and NMI are based on two different rationales. While ARI is related to pair counting measures, which are calculated based on the cluster and class membership of pairs of data points agreement; NMI is related to information-theoretic measures, which are based on entropic measures from information theory [21]. For all the clustering methods, we tested both Pearson and Euclidean distances for building the measure used as input to the respective clustering method. Finally, the result that we report in each table for each dataset is the best result considering the most effective combination of normalization and distance options. For best we mean the result that offers the highest values according to a mean rank taking into account accuracy, ARI and NMI.
.

### 2.4 Dataset description

Four different high-dimensional and nonlinear datasets were analysed in order to perform a comparative analysis of the clustering methods.

*Gastric mucosa microbiome dataset*
The dataset was generated by Paroni Sterbini and colleagues [22] and it consists of 24 biopsy specimens of the gastric antrum from 24 individuals who were referred to the Department of Gastroenterology of Gemelli Hospital (Rome) with dyspepsia symptoms (i.e. heartburn, nausea, epigastric pain and discomfort, bloating, and regurgitation). Twelve of these individuals (PPI1 to PPI12) had been taking PPIs for at least 12 months, while the others (S1 to S12) were not being treated (naïve) or had stopped treatment at least 12 months before sample collection. In addition, 9 (5 treated and 4 untreated) were positive for H. pylori infection, where H. pylori positivity (HP+) or negativity (HP-) was determined by histology and rapid urease tests. The number of features is 187 and indicates different microbial abundance. The metagenomics sequence data were processed, replicating the bioinformatics workflow followed by Paroni Sterbini [22], in order to obtain the dataset for the clustering algorithms. This dataset was analyzed for three clusters: HP+ (n=5), HP- (n=7) and PPI (n=12). The PPI patients with and without the presence of H. pylori are considered a unique class, because it is known from previous studies [23], [24] that PPI significantly changes the gastric environment and cover the effect of other factors such as HP presence. The data is public available in the NCBI Sequence Read Archive (SRA) (http://www.ncbi.nlm.nih.gov/sra, accession number SRP060417).

*Radar Signal dataset*
The data is composed by 350 radar signals targeting free electrons in the ionosphere, where each radar signal consisted of 34 features that are measurements of electromagnetic pulses. It was collected by the Space Physics Group of the Johns Hopkins University Applied Physics Laboratory [25]. The two groups are defined as: (1) 225 good radar signals, characterized by those signals that returned evidence of free electrons in the ionosphere, and (2) 125 bad radar signals which were those signals that passed through the ionosphere and returned background noise. Hence, good radar signals are similar, and bad radar signals might be dissimilar.
In the study of Cannistraci et al. [10] it was spotted that actually the bad radar signals might be segregated into two different groups. Therefore, here the data are analysed for both two and three clusters.

*The Tripartite-Swiss-Roll dataset*
In order to 'objectively' (using a ground truth) test how the clustering algorithms could detect nonlinear relationships, we additionally performed an analysis on the Tripartite-Swiss-Roll dataset (Fig.1): an artificial dataset characterized by evident nonlinear patterns and generated as discretization of the manifold associated to a Swiss-Roll function [26] in a three-dimensional (3D) space. Indeed, it is a synthetic dataset composed by 723 points obtained as the partition in three sections of a discrete Swiss-Roll manifold depicted in a three-dimensional space [26]. It reproduces the typical nonlinearity (given by the Swiss-Roll shape) and the discontinuity (given by the tripartition of the manifold, and therefore three clusters), that might be often hidden in the multidimensional representation of data samples. However, we have to clarify that this dataset, contrarily to all the other ones used in this study, has significantly less features than sample, therefore it cannot be considered a multidimensional dataset. Yet, it is a very useful benchmark for nonlinear clustering.

*MNIST dataset*

MNIST [27] is one of the most used dataset in the machine learning. This is a large dataset that consists of 28x28 pixel images of handwritten digits. Every image can be thought as a 784-dimensional array, where each value represent each pixel's intensity in gray scale. The different sample groups are numbers between 0 and 9, for a total of 10 clusters. Since this is a very large dataset (60.000 samples), we randomly selected 300 samples for each digit, resulting in a sub-dataset with 3000 samples. Therefore, this dataset is composed by a total of 3000 samples, 784 features and 10 groups.

## 3 RESULTS AND DISCUSSION

The different clustering algorithms were compared and evaluated in four datasets (Gastric mucosa microbiome, Radar [2 clusters and 3 clusters], Tripartite-Swiss-Roll and MNIST) and their performance were determined using three measures commonly applied in partitioning tasks: Accuracy, ARI and NMI. In general, all performance measures are in concordance with respect to the best method, nevertheless, to be fair and provide a robust assessment, we declare the best method as the one that obtains the lowest value in average ranking between the three measures (mean rank). Therefore, the results reported in the tables are ordered according to mean rank.

In the case of the first real dataset, gastric mucosa microbiome (Table 2), all MC-kernel variations clearly improve the performance of MCL, particularly for MC-MCLl and MC-MCLo, moving MCL from the last place to leading positions. DBSCAN, which is ranked second, seems to work pretty well according to NMI and ARI, nevertheless, its accuracy is the lowest (0.58) among all methods. K-means, MCL, AP and MC-AP have the lowest performance with accuracy of ~0.67, ARI of ~0.2 and NMI between 0.21 and 2.6. In this first dataset we can conclude that the performance of the MC-MCL improves remarkably on all indicators in comparison to the other methods.

For the second real dataset that is composed of Radar signals - when two clusters are expected (Table 3) - K-means slightly outperforms the other clustering methods with accuracy of 0.71 and ARI and NMI of 0.18 and 0.14 respectively. All MC kernel variations improve consistently the linear MCL on the three performance measures, remarking again MC-MCLl, with values close to K-means performances (0.71, 0.17 and 0.12 of accuracy, ARI and NMI respectively). DBSCAN performance is also competitive compared with other methods and obtain the best NMI for this data together with K-means, while AP and MC-AP does not perform well in all three measures (maybe indicating a general issue with the type of clustering strategy). In this second dataset we can conclude that the performance of MC-MCL is in the same range and therefore comparable to the performance of K-means, which arrive first in the ranking. However, as mentioned in the dataset description section, the presence of a three cluster structure in this radar dataset was discovered by means of unsupervised dimension reduction in the study of Cannistraci et al., 2013 [10]. Indeed, it seems that the dualistic hypothesis of mere segregation in two groups composed of good and bad radar signals is too course and simplistic, and the bad signals clearly show a pattern of further segregation in two sub-groups of bad signals. Under this scenario, if we label the data according to these three clusters, all methods seems to be negatively affected in accuracy by this new grouping (Table 4), with exception of the MC-MCL. In particular, MC-MCLo arrives first and its improvement in comparison to the not MC-MCL methods is significant. An important note is the possibility to comment what is the adequate cluster structure to consider in this dataset. In general, nonlinear clustering with hypothesis of three clusters achieves better results (MC-MCLo reaches accuracy, NMI and ARI respectively of 0.74, 0.27 and 0.38) than linear clustering with hypothesis of two clusters that achieves visible lower results (K-means reaches respectively 0.71, 0.18 and 0.14). In this second real dataset we can conclude that the hypothesis of three clusters seems perhaps more likely than two clusters, and that the presence of data nonlinearity might be an obstacle that in previous studies hindered this conclusion. But, with the help on nonlinear clustering such as MC-MCL such types of 'difficult' data can be approached, and can reveal some hidden aspects of their nonlinear structure.

The third is a synthetic dataset, which we term Tripartite-Swiss-Role (see Fig.1 and details in the respective data description section above) - that we adopt to offer a didactic example of how nonlinear clusters appears in a 3D space. The results in Table 5 shows that MCL, MC-MCLo, MC-MCLl and DBSCAN find perfectly the clusters of the Tripartite-Swiss-Roll. Surprisingly, MC-MCLs is not able to find the correct number of clusters and therefore obtain an accuracy of 0, but according to ARI and NMI values (0.47 and 0.63 respectively) it is still able to perform as MC-AP. AP and K-means are linear clustering methods and, as expected according to theory, on a nonlinear clustering problem results the lowest in performance. A possible explanation for the different result of this MC-MCL variant normalized with square root is due to the deformation of distances with this normalization. While long distances are shrunk, short distances (lower than one) are actually a bit amplified. This trend does not occur with the log(1+x) normalization used on the MC-distance-kernel. This third synthetic dataset is the only case in the present study where, in presence of a nonlinear clustering structure, classical MCL can achieve comparable performance to MC-MCL. Indeed, in all the three real datasets previously analysed, MCL was the worst algorithm between the 5 different types tested. This findings on one side suggest the utility to adopt synthetic data because yet on this example linear clustering algorithms such as AP and K-means result, as theoretical expected, the worst. On the other side, the same results suggest that simple synthetic datasets with many samples and few dimensions, although are an interesting and useful benchmark, might be too 'naïvely' designed. They might miss other crucial aspects of data nonlinearity which emerge in case of curse of dimensionality (when the number of features is substantially larger than the number of samples). Altogether, after this didactic example we can conclude that it is important to expose the tested algorithms to different data scenarios in which nonlinearity emerges from different data sources. Till now we tested unsupervised recognition of nonlinear patterns that emerge from genomic, radar signal and synthetic backgrounds, but we considered always the scenario of few number of expected clusters. It is now time to confront these algorithms on a more challenging benchmark, which contains 10 hidden clusters and is one of the most famed (or maybe 'ill-famed' for its difficulty) benchmarks for testing nonlinear pattern recognition performance.

The forth considered dataset is the well-known MNIST, which contains grayscale images of handwritten one-cipher numbers from zero to nine. From Table 6 and Fig.2 emerges patently that all MC variants (both MC-MCL and MC-AP - however also here as in the rest of the article MC-MCL overcomes MC-AP) perform significantly better (pvalue<0.05, Mann-Whitney test used) than their linear variants and the rest of the methods. Indeed the MC-MCL variants achieve values of more than 0.7 in accuracy. In particular, from Fig.2 emerges that the average increment of MC-based methods on not MC-based is 59% in accuracy, 140% in ARI and 43% in NMI. This is clearly an outstanding result considering on one side the simplicity of the MC nonlinear strategy and the fact that is parameter-free, on the other side the intrinsic difficulty to reach 'high' unsupervised pattern recognition performance on this nonlinear dataset. Of course, the term 'high' to evaluate the performance of MC methods should be considered with a 'grain of salt', as a comparison to classical clustering methods whose performance is significantly lower and around 0.5 accuracy. In particular, K-means performs poorly with 0.57 of accuracy, while MCL does not surpass a value 0.5 and DBSCAN of 0.3 for the same measurement. Notice the difficulty of this dataset with 10 different clusters and many similar handwritten digits between clusters.

A summary of accuracies across all datasets is presented in Table 7. The methods are ordered by mean accuracy performance. It is clearly appreciated that in general MC-MCL (with exception of MC-MCLs in the Tripartite-Swiss-roll data, where it is not able to find the correct number of clusters obtaining an accuracy equal to zero) improves the performance of classical MCL in all datasets and turns MCL in one of the best clustering methods for nonlinear data among the compared algorithms. Similarly, MC-AP improves in general the performance of AP on these nonlinear data, and since it is the second best methodology and adopt also the MC strategy, we can conclude that in general MC seems to offer an effective and promising framework to improve nonlinear pattern recognition in future studies. DBSCAN is a well-known clustering algorithm, which still performs competitively. AP and K-means performance is lower with respect to the rest of the methods and this is in agreement with the expectations since they are linear clustering methods.

Evaluating the methods according to mean accuracy across data is not always as informative as an evaluation by mean ranking. Indeed, in Table 8 the magnitude of the accuracy does not affect directly the final average as it occurs in Table 7 for MC-MCLs. In Table 8, it is clear that MC strategy improves the performance of MC-MCL for all its versions; MC-AP seems to work better as well than its linear version; K-means and DBSCAN perform close to one another except for the radar dataset, where K-means was the strongest method whereas DBSCAN one of the weakest. Finally, MCL obtains the worst average ranking. This last finding on one side is at support of the fact that the average ranking is a proper way to summarize the results, since MCL is a linear method whose performance was the worst for the majority of the tests; on the other side advocates the importance to design adequate nonlinear variations of effective linear methods such as MCL, in order to allow a 'deep' analysis of big data and their hidden patterns.

Finally, we suggest to carefully explore in all studies different types of linear and nonlinear pattern recognition algorithms. This can offer a better understanding of the data and can avoid to reach misleading conclusions that could arise by using only one of the strategies.

## ACKNOWLEDGMENT

C.D. is funded by the Research Grants – Doctoral Programs in Germany (DAAD), Promotion program Nr: 57299294. C.V.C is the corresponding author.

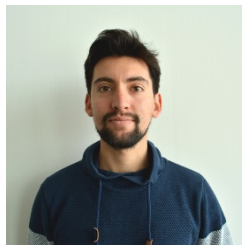

**Claudio Durán** received the Engineer diploma degree in Bioinformatics from the University of Talca, Chile, in 2016. He is currently a computer science Ph.D. student in the Biomedical Cybernetics Group led by Dr. Carlo Vittorio Cannistraci at the Biotechnology Center of the Technische Universität Dresden, Germany. His research interests include machine learning, network science, and systems biomedicine.

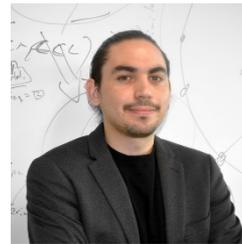

**Aldo Acevedo** received the Engineer diploma degree in Bioinformatics from the University of Talca, Chile, in 2015. He worked as full-stack developer in the Social Responsibility Department at the University of Talca from 2015 until 2016. He is currently a computer science Ph.D. student in the Biomedical Cybernetics Group led by Dr. Carlo Vittorio Cannistraci at the Biotechnology Center of the Technische Universität Dresden, Germany. His research interests include software engineering, artificial intelligence, and bioinformatics.

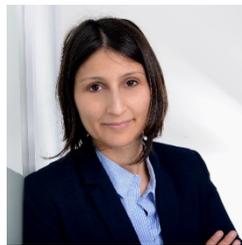

**Sara Ciucci** received the B.Sc. degree in Mathematics from the University of Padova, Italy, in 2010 and the M.Sc. degree in Mathematics from the University of Trento, Italy in 2014. She obtained her Ph.D. degree in Physics from Technische Universität Dresden, Germany, in 2018 under the supervision of Dr. Carlo Vittorio Cannistraci in the Biomedical Cybernetics Group, where she remained as postdoctoral researcher till March 2019. Her research interests include machine learning, network science, and systems biomedicine.

**Alessandro Muscoloni** was born in Italy in 1992. He received the B.S. degree in Computer Engineering (2014), the M.S. degree in Bioinformatics (2016) from the University of Bologna, Italy and the PhD in Computer Science and Engineering (2019) from the Technical University of Dresden, Germany. He is currently a Post doctoral researcher under the supervision of Dr. Carlo Vittorio Cannistraci in the Biomedical Cybernetics Lab at the Biotechnology Center (BIOTEC) of the TU-Dresden, Germany.

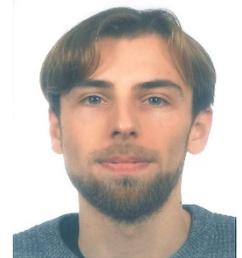

His research interests include complex network and machine learning. He was awarded a scholarship to meritorious students enrolled at the University of Bologna in the AY 2015/2016, and he was selected by the Graduate Academy of TU-Dresden as awardee of the Saxon Scholarship Program (Sept. 2016 – Aug. 2017) and twice of the Travel Award for Conferences (Dec. 2016, Dec. 2017). He is a member of the Network Science Society.

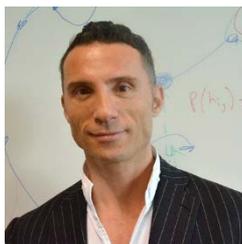

**Carlo Vittorio Cannistraci** is a theoretical engineer and was born in Milazzo, Sicily, Italy, in 1976. He received the M.S. degree in Biomedical Engineering from the Polytechnic of Milano, Italy, in 2005 and the Ph.D. degree in Biomedical Engineering from the Interpolytechnic School of Doctorate, Italy, in 2010. From 2009 to 2010, he was visiting scholar in the Integrative Systems Biology lab of Dr. Trey Ideker at the University California San Diego (UCSD), USA. From 2010 to 2013, he was postdoc and then research scientist in machine intelligence and complex network science for personalized biomedicine at the King Abdullah University of Science and Technology (KAUST), Saudi Arabia. Since 2014, he has been Independent Group Leader and Head of the Biomedical Cy-bernetics lab at the Biotechnological Center (BIOTEC) of the TU-Dresden, Germany. He is also affiliated with the MPI Center for Systems Biology Dresden and with the Tsinghua Laboratory of Brain and Intelligence (China). He is author of three book chapters and more than 40 articles. His research interests include subjects at the interface between physics of complex systems, complex networks and machine learning, with particular interest for applications in biomedicine and neuroscience. Dr. Cannistraci is member of the Network Science Society, member of the International Society in Computational Biology, member of the American Heart Association, member of the Functional Annotation of the Mammalian Genome Consortium. He is Editor for the mathematical physics board of the journal Scientific Reports (edited by Nature) and of PLOS ONE. *Nature Biotechnology* selected his article (*Cell* 2010) on machine learning in developmental biology to be nominated in the list of 2010 notable breakthroughs in computational biology. *Circulation Research* featured his work (*Circulation Research* 2012) on leveraging a cardiovascular systems biology strategy to predict future outcomes in heart attacks, commenting: "a space-aged evaluation using computational biology". TU Dresden honoured Dr. Cannistraci of the *Young Investigator Award 2016 in Physics* for his recent work on the local-community-paradigm theory and link prediction in monopartite and bipartite complex networks. In 2017, Springer-Nature scientific blog highlighted with an interview to Dr. Cannistraci his recent study on "How the brain handles pain through the lens of network science". In 2018, the American Heart Association covered on its website Dr. Cannistraci's chronobiology discovery on how the sunshine affects the risk and time onset of heart attack. In 2018, Nature Communications featured Carlo's article entitled "Machine learning meets complex networks via coalescent embedding in the hyperbolic space" in the selected interdisciplinary collection of recent research on complex systems.

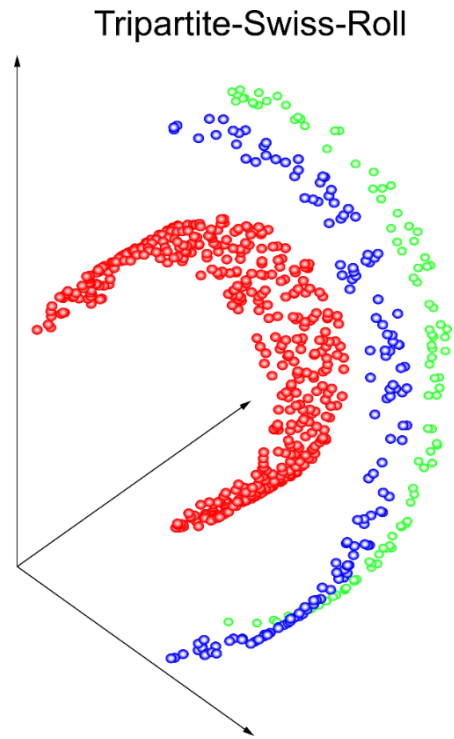

**Figure 1. Tripartite-swiss-Roll.**
Visualization in a 3-Dimensional space of the non-linear manifold of the synthetic dataset Tripartite-Swiss Roll and its respective three clusters (red, blue and green).

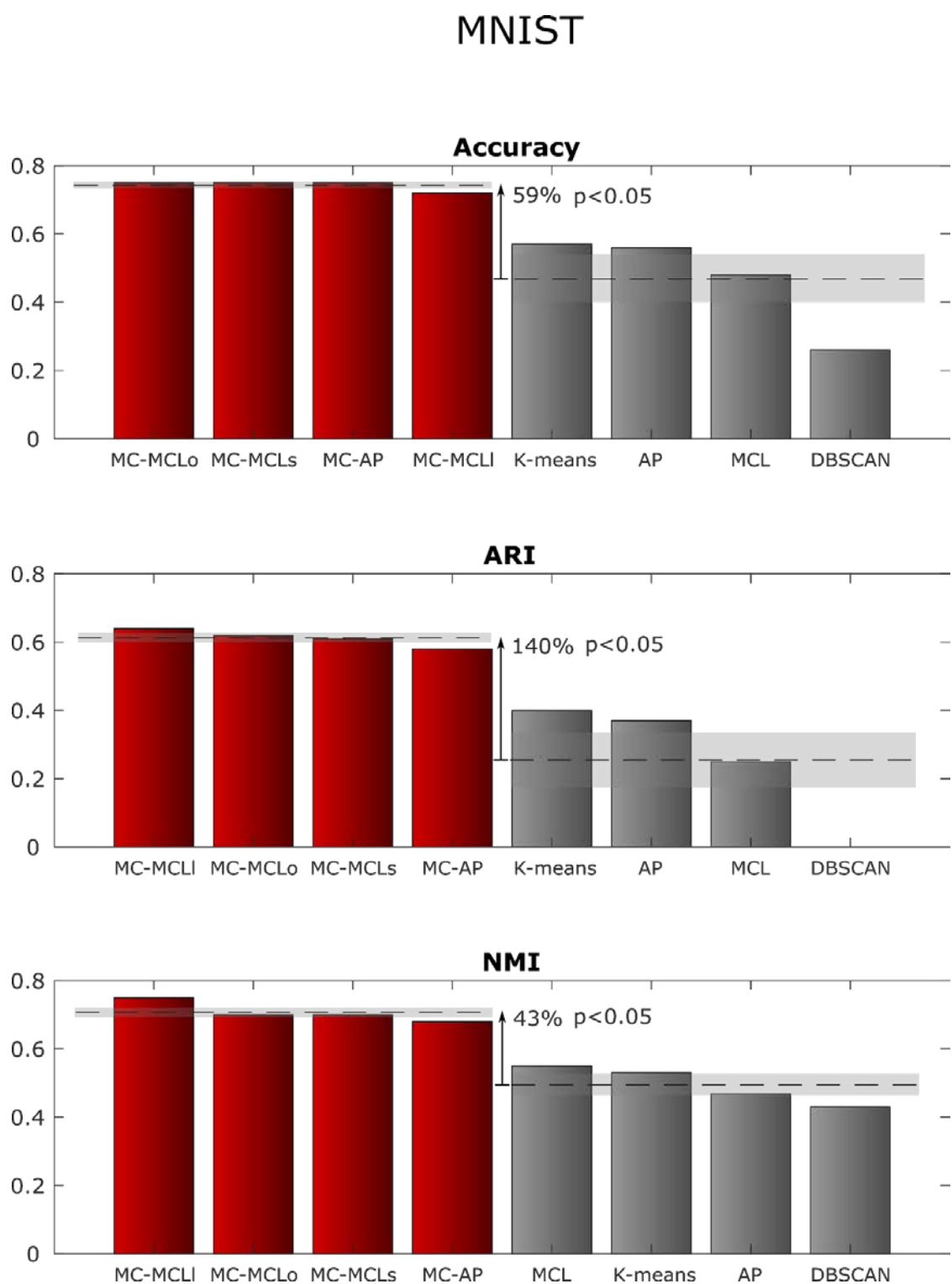

**Figure 2.  Performance comparison in MNIST**.
Accuracy, ARI and NMI performance comparison of the MC-methods (red) against the non-MC methods (gray) in the MNIST dataset. In dashed line the respective group (MC and non-MC) means. The arrows represent the percentage of improvement from the MC and non-MC group means. P represents the Mann-Whitney p-value.

**Table 1. Pseudocode of the MC-MCL algorithm with the respective input and output arguments**

| |
|---|
| **Inputs:** |
| **x,** $n \times m$ numeric data matrix ($n$ = number of samples; $m$ = number of features); |
| **C,** numeric value determining the **number** of clusters to find; |
| **dist,** specifies the distance applied on 'x' to construct the graph; |
| **factor,** specifies different MC-MCL variants (options: original, squared root or logarithm); |
| **Outputs:** |
|    **clust,** numeric vector of $n \times 1$ containing the clustering result of MC-MCL; |
| **Pseudocode:** |
|    D = distance_construction(x, dist); |
|    If factor == 1, no changes in D; end if |
|    If factor == 2, D = square root of D; end if |
|    If factor == 3, D = natural logarithm of (1 + D); end if |
|    T = extract the minimum spanning tree out of D; |
|    MC_distance = compute distances between all node pairs over T; |
|    If dist is Euclidean: |
|      MC_distance = each element of MC_distance is divided by the maximum value of MC_distance; |
|    end if |
|    S = 1 − MC_distance; |
|    S = enforcing_sparcity(S); |
|    clust = compute_mcl(S,C); |
|    returning clust as output; |
|   |
|    function enforcing_sparcity(Input: S): |
|      unique_weights = unique values of S; |
|      idx = indices of the positive values of unique_weights ordered from the lowest to the highest value; |
|      for i = 1… length of idx: |
|        cutoff = unique_weights(idx(i)); |
|        temporal_S = S; |
|        elements of temporal_S that are lower than the cutoff are set to 0; |
|        elements that are equal or greater than the cutoff in temporal_S are subtracted by the cutoff; |

```
        number_Components = check number of components of temporal_S;
        if number_Components > 1:
            if i == 1:
                warning lunched: for the first cutoff the number_Components is already larger than 1;
                break the for loop;
            else:
                cutoff = select previous cutoff as unique_weights(idx(i - 1));
                break for loop;
            end if-else
        end if
    end for loop
    elements of S that are lower than the cutoff are set to 0;
    elements that are equal or greater than the cutoff in S are subtracted by the cutoff;
    returning S as output of the enforcing_sparcity function;

function compute_mcl(Input: S and C):
    MCL_file_input = creating txt input file needed for MCL containing the network in S;
    min_inflation = 1.1;
    max_inflation = 20;
    resolution = [0.1, 0.01 and 0.001];
    comment: "initializing variables for binary search"
    left = 1;
    inflation = vector from min_inflation until max_inflation in steps of resolution(1);
    right = length of inflation;
    idx = integer of the middle point between left and right;
    number_clusters = 0;
    for res = each element of resolution:
        comment: "section to lower the resolution if needed"
        if left > right:
            if last_left == last_right:
                if number_clusters > C:
                    last_left = last_left − 1;
                else:
```

```
                last_right = last_right + 1;
            end if-else
            if last_right > than the length of inflation or if last_left < 1:
                break the for loop because out of inflation bounds [1.1, 20];
            end if
        end if
        inflation = vector from inflation(last_left) until inflation(last_right) in steps of res
        left = 1;
        right = length of inflation;
        idx = integer of the middle point between left and right;
    end if
    comment: "computing MCL with binary search for inflation"
    while number_clusters is different from C:
        MCL_file_output = compute MCL(MCL_file_input, inflation(idx));
        clusters = obtain clusters from MCL_file_output;
        number_clusters = obtain number of clusters from clusters;
        last_left = left;
        last_right = right;
        if number_clusters > C:
            right = idx − 1;
        else if number_clusters < C:
            left = idx + 1;
        end if-else
        if left > right, break while loop; end if
        idx = integer of the middle point between left and right;
    end while
    if number_clusters == C, break for loop; end if
 when the correct number of clusters is found or the for loop arrives until the end, the clusters
variable is returned as output of the compute_mcl function
```

**Table 2. Clustering performance in Gastric mucosa microbiome data.** Accuracy (Acc), Adjusted Rand Index (ARI), Normalized Mutual Information (NMI), and mean rank (according to the previous mentioned measures) are reported for each clustering method together with the best distance approach (Pearson correlation or Euclidean) and normalization (Norm) applied. The methods are sorted by mean rank from the highest (top) to the lowest (bottom) performance.

| Methods | Best distance | Norm | Acc | ARI | NMI | Mean Rank |
|---|---|---|---|---|---|---|
| MC-MCLl | corr | LOG | 0.71 | 0.29 | 0.31 | 1.3 |
| MC-MCLo | corr | LOG | 0.71 | 0.29 | 0.31 | 1.3 |
| DBSCAN | corr | - | 0.58 | 0.28 | 0.38 | 2.0 |
| MC-MCLs | corr | LOG | 0.71 | 0.26 | 0.31 | 2.0 |
| K-means | corr | LOG | 0.67 | 0.20 | 0.26 | 3.0 |
| AP | corr | LOG | 0.67 | 0.20 | 0.24 | 3.3 |
| MC-AP | corr | LOG | 0.67 | 0.20 | 0.24 | 3.3 |
| MCL | corr | LOG | 0.67 | 0.19 | 0.21 | 4.0 |

**Table 3. Clustering performance in Radar (two clusters) data.** Accuracy (Acc), Adjusted Rand Index (ARI), Normalized Mutual Information (NMI) and mean rank (according to the previous mentioned measures) are reported for each clustering method together with the best distance approach (Pearson correlation or Euclidean) and normalization (Norm) applied. The methods are sorted by mean rank from the highest (top) to the lowest (bottom) performance.

| Methods | Best distance | Norm | Acc | ARI | NMI | Mean Rank |
|---|---|---|---|---|---|---|
| K-means | eucl | - | 0.71 | 0.18 | 0.14 | 1.0 |
| AP | eucl | - | 0.71 | 0.17 | 0.13 | 1.7 |
| MC-MCLl | eucl | - | 0.71 | 0.17 | 0.12 | 2.0 |
| DBSCAN | corr | - | 0.68 | 0.11 | 0.14 | 2.7 |
| MC-MCLs | corr | - | 0.69 | 0.09 | 0.13 | 3.0 |
| MC-AP | eucl | - | 0.69 | 0.14 | 0.09 | 3.3 |
| MC-MCLo | corr | - | 0.68 | 0.07 | 0.11 | 4.3 |
| MCL | eucl | - | 0.60 | 0.04 | 0.06 | 5.7 |

**Table 4. Clustering performance in Radar (three clusters) data.** Accuracy (Acc), Adjusted Rand Index (ARI), Normalized Mutual Information (NMI) and mean rank (according to the previous mentioned measures) are reported for each clustering method together with the best distance approach (Pearson correlation or Euclidean) and normalization (Norm) applied. The methods are sorted by mean rank from the highest (top) to the lowest (bottom) performance.

| Methods | Best distance | Norm | Acc | ARI | NMI | Mean Rank |
|---|---|---|---|---|---|---|
| MC-MCLs | corr | - | 0.74 | 0.27 | 0.38 | 1.0 |
| MC-MCLl | corr | - | 0.70 | 0.20 | 0.35 | 2.3 |
| MC-MCLo | corr | - | 0.70 | 0.18 | 0.32 | 3.0 |
| AP | eucl | - | 0.66 | 0.23 | 0.26 | 3.0 |
| K-means | eucl | - | 0.62 | 0.16 | 0.15 | 5.3 |
| MC-AP | corr | - | 0.56 | 0.08 | 0.22 | 5.7 |
| DBSCAN | eucl | - | 0.64 | 0.01 | 0.02 | 6.7 |
| MCL | eucl | - | 0.43 | 0.03 | 0.05 | 7.0 |

**Table 5. Clustering performance in Tripartite-Swiss-Roll data.** Accuracy (Acc), Adjusted Rand Index (ARI), Normalized Mutual Information (NMI) and mean rank (according to the previous mentioned measures) are reported for each clustering method together with the best distance approach (Pearson correlation or Euclidean) and normalization (Norm) applied. The methods are sorted by mean rank from the highest (top) to the lowest (bottom) performance. *Only case in all comparisons of this study that the algorithm cannot detect the correct number of clusters.

| Methods | Best distance | Norm | Acc | ARI | NMI | Mean Rank |
|---|---|---|---|---|---|---|
| MC-MCLl | eucl | - | 1.00 | 1.00 | 1.00 | 1.0 |
| MC-MCLo | eucl | - | 1.00 | 1.00 | 1.00 | 1.0 |
| DBSCAN | eucl | - | 1.00 | 1.00 | 1.00 | 1.0 |
| MCL | eucl | - | 1.00 | 1.00 | 1.00 | 1.0 |
| MC-AP | eucl | - | 0.64 | 0.47 | 0.59 | 2.3 |
| MC-MCLs | eucl | - | 0.00* | 0.47 | 0.63 | 3.0 |
| K-means | eucl | - | 0.56 | 0.10 | 0.21 | 3.3 |
| AP | eucl | - | 0.54 | 0.09 | 0.19 | 4.3 |

**Table 6. Clustering performance in MNIST data.** Accuracy (Acc), Adjusted Rand Index (ARI), Normalized Mutual Information (NMI) and mean rank (according to the previous mentioned measures) are reported for each clustering method together with the best distance approach (Pearson correlation or Euclidean) and normalization (Norm) applied. The methods are sorted by mean rank from the highest (top) to the lowest (bottom) performance.

| Methods | Best distance | Norm | Acc | ARI | NMI | Mean Rank |
|---|---|---|---|---|---|---|
| MC-MCLl | corr | - | 0.72 | 0.64 | 0.75 | 1.3 |
| MC-MCLo | eucl | LOG | 0.75 | 0.62 | 0.70 | 1.7 |
| MC-MCLs | eucl | LOG | 0.75 | 0.61 | 0.70 | 2.0 |
| MC-AP | eucl | LOG | 0.75 | 0.58 | 0.68 | 2.7 |
| K-means | corr | LOG | 0.57 | 0.40 | 0.53 | 4.3 |
| AP | eucl | LOG | 0.56 | 0.37 | 0.47 | 5.3 |
| MCL | corr | LOG | 0.48 | 0.25 | 0.55 | 5.3 |
| DBSCAN | corr | LOG | 0.26 | 0.00 | 0.43 | 7.0 |

**Table 7. Mean accuracy in clustering performance across all data.** The table reports, for each clustering algorithm, the accuracy for all the datasets and the mean accuracy (Mean Acc) over the datasets. The methods are sorted by mean accuracy from the highest (top) to the lowest (bottom) performance.

| Methods | Gastric mucosa | Swiss-Roll | Radar | Radar(3C) | MNIST | Mean Acc |
|---|---|---|---|---|---|---|
| MC-MCLl | 0.71 | 1.00 | 0.71 | 0.7 | 0.72 | 0.77 |
| MC-MCLo | 0.71 | 1.00 | 0.68 | 0.7 | 0.75 | 0.77 |
| MC-AP | 0.67 | 0.64 | 0.69 | 0.56 | 0.75 | 0.66 |
| MCL | 0.67 | 1.00 | 0.6 | 0.43 | 0.48 | 0.64 |
| DBSCAN | 0.58 | 1.00 | 0.68 | 0.64 | 0.26 | 0.63 |
| AP | 0.67 | 0.54 | 0.71 | 0.66 | 0.56 | 0.63 |
| K-means | 0.67 | 0.56 | 0.71 | 0.62 | 0.57 | 0.63 |
| MC-MCLs | 0.71 | 0.00 | 0.69 | 0.74 | 0.75 | 0.58 |

**Table 8. Mean rank in clustering performance across all data.** The table reports, for each clustering algorithm, the ranking according to accuracy for all the datasets and the mean rank over the datasets. The methods are sorted by mean rank from the highest (top) to the lowest (bottom) rank.

| Methods | Gastric mucosa | Swiss-Roll | Radar | Radar(3C) | MNIST | Mean |
|---|---|---|---|---|---|---|
| MC-MCLl | 1 | 1 | 1 | 2 | 2 | 1.4 |
| MC-MCLo | 1 | 1 | 3 | 2 | 1 | 1.6 |
| MC-MCLs | 1 | 5 | 2 | 1 | 1 | 2.0 |
| MC-AP | 2 | 2 | 2 | 6 | 1 | 2.6 |
| AP | 2 | 4 | 1 | 3 | 4 | 2.8 |
| K-means | 2 | 3 | 1 | 5 | 3 | 2.8 |
| DBSCAN | 3 | 1 | 3 | 4 | 6 | 3.4 |
| MCL | 2 | 1 | 4 | 7 | 5 | 3.8 |